\pdfoutput=1

\documentclass[11pt]{article}

\usepackage[preprint]{coling}
\usepackage{comment}
\usepackage{times}
\usepackage{latexsym}
\usepackage{rotating}
\usepackage{booktabs}
\usepackage{amsfonts}
\usepackage{float}
\PassOptionsToPackage{a4paper,margin=1in}{geometry}
\usepackage[a4paper,margin=1in]{geometry} 

\usepackage[T2A,T1]{fontenc}
\DeclareRobustCommand{\cyr}[1]{%
  {\fontencoding{T2A}\selectfont#1}%
}
\usepackage[ukrainian,english]{babel}

\usepackage{diagbox} 
\usepackage{booktabs} 
\usepackage{graphicx} 
\usepackage[utf8]{inputenc}
\usepackage{microtype}
\usepackage{inconsolata}
\usepackage{graphicx}
\usepackage{amsmath} 

\title{Benchmarking Multimodal Models for Ukrainian Language Understanding Across Academic and Cultural Domains}

\author{Yurii Paniv \\
  Ukrainian Catholic University \\
  \small{\texttt{paniv@ucu.edu.ua}} \\\And
  Artur Kiulian \\
  OpenBabylon \\
  \small{\texttt{akiulian@gmail.com}} \\\And
  Dmytro Chaplynskyi \\
  lang-uk initiative \\
  \small{\texttt{chaplinsky.dmitry@gmail.com}} \\\And
  Mykola Khandoga \\
  OpenBabylon \\
  \small{\texttt{mkhandoga@gmail.com}} \\\AND
  Anton Polishko \\
  OpenBabylon \\
  \small{\texttt{anton.polishko@gmail.com}} \\\And
  Tetiana Bas \\
  Minerva University \\
  \small{\texttt{tetiana@uni.minerva.edu}} \\\And
  Guillermo Gabrielli \\
  OpenBabylon \\
  \small{\texttt{guillermo.gabrielli.fer@gmail.com}} \\}

\begin{document}
\maketitle
\begin{abstract}
While the evaluation of multimodal English-centric models is an active area of research with numerous benchmarks, there is a profound lack of benchmarks or evaluation suites for low- and mid-resource languages. We introduce ZNO-Vision, a comprehensive multimodal Ukrainian-centric benchmark derived from standardized university entrance examination (ZNO). The benchmark consists of over 4,300 expert-crafted questions spanning 12 academic disciplines, including mathematics, physics, chemistry, and humanities. We evaluated the performance of both open-source models and API providers, finding that only a handful of models performed above baseline. Alongside the new benchmark, we performed the first evaluation study of multimodal text generation for the Ukrainian language: we measured caption generation quality on the Multi30K-UK dataset, translated the VQA benchmark into Ukrainian, and measured performance degradation relative to original English versions. Lastly, we tested a few models from a cultural perspective on knowledge of national cuisine. We believe our work will advance multimodal generation capabilities for the Ukrainian language and our approach could be useful for other low-resource languages.
\end{abstract}

\section{Introduction}

Vision-language models (VLMs) have expanded LLM capabilities into more domains, allowing for models to work with plenty of new tasks such as OCR \cite{liu2024ocrbenchhiddenmysteryocr}, image captioning, visual question answering and many more. 

While numerous benchmarks~\cite{mmlm_bench_survey} evaluate VLMs performance across a range of multimodal tasks, these resources primarily serve English-language models, underscoring a critical gap for evaluating VLMs in less-resourced languages. This absence is especially pronounced for Ukrainian, where multimodal benchmarks are exceedingly scarce. 

Our work addresses this gap by introducing a suite of Ukrainian-specific benchmarks and presenting benchmarking results for leading proprietary and open-source VLMs. To estimate academic knowledge, we developed a new benchmark based on the External Independent Evaluation (ZNO) - national university entrance exam
\cite{eie-main-page}, which includes a large selection of questions across various fields, such as chemistry, mathematics, Ukrainian language and literature, etc.
Besides that, we evaluated all models using Multi30K-UK \cite{saichyshyna-etal-2023-extension}, one of the few existing Ukrainian multimodal benchmarks. Also, we translated the VQA benchmark \cite{vqa} and measured the performance degradation compared to the English version. Additionaly, for culture test, we developed a new multimodal benchmark, UACUISINE, based on 20 popular Ukrainian dishes and finetuned Paligemma model on UACUISINE training dataset, improving performance on UACUISINE benchmark and VQA.

We believe that our effort would advance the development of VLMs applications for the Ukrainian language across academic and business sectors worldwide, wherever it's being used.

\section{Related Work}
\label{sec:related-work}
Recent years have seen significant development in multimodal benchmarks for evaluating VLMs. Existing benchmarks can be broadly categorized into three groups.
General visual understanding benchmarks include VQA~\cite{vqa} (1M+ question-answer pairs), GQA~\cite{gqa} (compositional reasoning), and MMMU~\cite{mmmu} (broad domain reasoning).
Cultural and multilingual benchmarks are represented by CulturalVQA~\cite{cultural} (11 countries), WorldCuisines~\cite{worldcusines} (30 languages), and MaXM~\cite{maxm} (7 languages).
Visual reasoning benchmarks feature CLEVR~\cite{clevr} (compositional reasoning), A-OKVQA~\cite{okvqa} (external knowledge), and Visual7W~\cite{visual7W} (semantic understanding).

While these benchmarks provide comprehensive evaluation frameworks, they predominantly focus on English language capabilities. Recent multilingual benchmarks often rely on translations rather than culturally-grounded content, highlighting a critical gap for evaluating VLMs in underrepresented languages like Ukrainian. Translation-based benchmarks like xGQA \cite{pfeiffer2022xgqacrosslingualvisualquestion} (9,670 questions in 7 languages) often introduce artifacts and fail to capture cultural nuances~\cite{translation_artifacts}. Current cultural evaluations are either too limited in scope (CulturalVQA: 2,378 questions across 11 countries) or too narrow in focus (WorldCuisines: food-specific across 30 languages).
Analyzing the WorldCuisines, we found three critical limitations regarding Ukrainian cuisine: (1) representation was restricted solely to location identification tasks without deeper cultural assessment, (2) the selection of dishes failed to capture the breadth of Ukrainian culinary traditions, and (3) several dishes were incorrectly categorized as Ukrainian while featuring Russian-language captions and representing Russian cuisine variants.

\section{Datasets \& Methodology}

\noindent\textbf{ZNO multi-choice questions.}
External Independent Evaluation (abbr. "ZNO" in Ukrainian) is a national Ukrainian test for high school graduates \cite{eie-main-page}. This test is challenging for LLMs even in text-only setting \cite{romanyshyn-etal-2024-unlp}. We gathered questions from Osvita portal ~\cite{Osvita_Test}, where an image is required for the answer. Dataset consists of 
4306 question-pair in 13 categories (overview in \autoref{sec:appendix_B}): Math, Geography, Ukrainian language and literature, 
Teaching, History, Spanish, German, French, English, Chemistry, Physics, Biology and Other (for a small portion of unclassified questions). From our source dataset, we filtered out questions with multiple images, images as answers, and choice-matching questions to streamline the setting for the benchmark, leaving only questions that require a single letter (e.g., B) as an answer.

\noindent\textbf{Multi30K-UK.} We evaluated models for caption generation task on Multi30K-UK \cite{saichyshyna-etal-2023-extension} dataset. We use Flickr2017 and Flickr2018 datasets as dev and test subsets respectively.

\noindent\textbf{Visual Question Answering (VQA) 2.0.}
We translated portions of an English-language benchmark VQA 2.0 \cite{vqa} to measure performance degradation for translated Ukrainian multimodal evaluation. We used the standard evaluation process for VQA \cite{VQA_Evaluation} on a subset of 1000 questions.

We used GPT-4o for translation due to its quality \cite{paniv-etal-2024-setting} and ability to work in long-context setting. We selected 1,000 question-answer pairs from VQA 2.0, preserving original associations and metadata.

\noindent\textbf{UACUISINE Benchmark.}
In this dataset we addressed the issues with WorldCuisine dataset mentioned in \autoref{sec:related-work}.
The UACUISINE benchmark consists of seven question types across three categories: (1) dish identification (three variants), (2) text generation (ingredients and recipe), and (3) characteristic classification (temperature and taste). The identification questions were adapted from WorldCuisines and translated into Ukrainian, while preparation and classification questions were newly introduced to assess deeper culinary understanding. We curated a dataset of 20 most typical Ukrainian dishes and annotated each with 7 question types in Ukrainian, generating 140 question-answer pairs.

\noindent\textbf{Evaluation Framework.}
For ZNO benchmark the model is given an image and a natural language question about the image. The expected answer is a letter, e.g. A/B/C/D. The dataset contains 491/490/3325 (dev/validation/test) samples, each making up an image, a question and multi-choice answers encoded as letters.

54\% of questions are pure visual questions to test OCR capabilities for models. We prepend questions with a single prompt: \cyr{"дай відповідь на питання і напиши варіант відповіді в квадратних дужках, наприклад: "[А]""}\latintext (answer this question and write the answer in quadratic braces, in example "[A]"). We extraxt the last match for quadratic braces using regular expression to replicate the original setting of the test: final answer could be counted only if it's strictly in correct box.

For the translated VQA 2.0 benchmark, we maintained the original evaluation protocols \cite{VQA_Evaluation} to ensure comparability with English-language results. Specifically, we used the VQA accuracy metric that takes into account answer agreement among multiple human annotators.

\begin{table}
\centering
\begin{tabular}{|l|c|c|}
\hline
\textbf{Model} & \textbf{ZNO val} & \textbf{ZNO test} \\ \hline
Gemini Pro & 0.680 & 0.675 \\ \hline
Claude 3.5 Sonnet & 0.651 & 0.643 \\ \hline
Qwen2-VL-72B & 0.496 & 0.512 \\ \hline
GPT-4o & 0.416 & 0.470 \\ \hline
Qwen2-VL-7B & 0.247 & 0.264 \\ \hline
Baseline & 0.224 & 0.219 \\ \hline
Llama-3.2-11B & 0.116 & 0.095 \\ \hline
LLaVa-v1.6-mistral-7b & 0.071 & 0.067 \\ \hline
Paligemma-3b  & 0.049 & 0.058 \\ \hline
Pixtral-12b  & 0.000 & 0.000 \\ \hline
\end{tabular}
\caption{Accuracy scores on ZNO dataset across different models for validation and test subdatasets. More detailed results could de found in \autoref{sec:appendix_A}.}
\label{tab:zno}
\end{table}

Evaluation for UACUISINE consists of three metrics. For the dish name prediction and characteristic classification we use exact match score (EM). For the ingredients generation, we use a matching score called the Intersection Match (IM) (detailed formulas for both of them in \autoref{sec:intersection-match}).

For recipe generation evaluation, we use BERT score \cite{zhang2020bertscoreevaluatingtextgeneration} using "bert-base-multilingual-cased" model \cite{DBLP:journals/corr/abs-1810-04805} to capture semantic similarity.

For Multi30K-UK, we use SacreBLEU \cite{post-2018-call} and the same BERT score as well.

\section{Experimental Setup}
For each benchmark evaluation, we used their specific metrics. For all generations, we used a temperature equal to 1 and a maximum amount of tokens equal to 300. We evaluated both proprietary and open-source multimodal language models to provide a comprehensive assessment of current capabilities on Ukrainian language tasks.
For baseline evaluation of ZNO we select the first choice in each question, getting a 22\% accuracy score.

We additionally evaluated a fine-tuned Paligemma specifically on our UACUISINE training dataset to assess the potential of domain adaptation. Our training data includes 4,615 image-question-answer triplets from UACUISINE training dataset. We finetuned Paligemma using AdamW optimizer with learning rate of 2e-5, batch size 4, gradient accumulation steps 4.

\section{Results \& Discussion}

\begin{table}
\centering
\begin{tabular}{|p{0.15\textwidth}|c|c|}
\hline
\textbf{Model} & \textbf{BLEU} & \textbf{BERT} \\ \hline
Qwen2-VL-72B & $4.46 \pm 3.82$ & $0.728 \pm 0.043$ \\ \hline
Paligemma-3b         & $4.36 \pm 3.92$ & $0.706 \pm 0.045$ \\ \hline
Qwen2-VL-7B          & $3.35 \pm 2.71$ & $0.706 \pm 0.039$ \\ \hline
Qwen2-VL-2B          & $2.08 \pm 1.62$ & $0.667 \pm 0.050$ \\ \hline
LLaVa-v1.6-mistral-7b & $1.35 \pm 1.24$ & $0.620 \pm 0.037$ \\ \hline
Pixtral-12b          & $0.39 \pm 0.38$ & $0.604 \pm 0.032$ \\ \hline
\end{tabular}
\caption{Average scores $\pm$ STD in BLEU and BERT on Multi30k-UA dataset. }
\label{tab:multi30k}
\end{table}

In \autoref{tab:zno}, Gemini 1.5 Pro ~\cite{gemini} and Claude 3.5 Sonnet ~\cite{claude} demonstrated the best results on ZNO benchmark, with Qwen2-VL-72B ~\cite{qwen} being the only open source model to meaningfully beat a baseline.
Surprisingly, GPT-4o ~\cite{gpt4o} performed worse than Qwen, and LLaMA 3.2 ~\cite{llama3}, Paligemma-3B-mix-224 ~\cite{paligemma} and Pixtral~\cite{pixtral} failed to even beat a baseline.
The detailed breakdown of models performance per question category is provided in \autoref{sec:appendix_A}. 

As shown in \autoref{tab:combined-uacuisine-results}, the UACUISINE benchmark reveals potential benefits of finetuning, showing that Paligemma can outmatch Qwen with additional finetuning. 

As shown in \autoref{tab:multi30k}, testing caption generation task on Multi30K-Uk dataset did not provide a way to make conclusions about model performance. The target domain is too different (model frequently used synonyms) or did not follow instructions to provide an answer in one sentence only.

\noindent\textbf{Performance degradation in Ukrainian.}
Parallel benchmarking in English and Ukrainian VQA datasets provides an insight into language-induced performance degradation as shown in \autoref{tab:vqa_results}. While all the models demonstrated a certain degree of performance degradation in Ukrainian, it varied substantially. We attribute performance degradation to translation model performance and assumption that models are more capable of understanding and reasoning in English rather than in Ukrainian. 
While the former factor is a limitation of our work, the latter is a manifestation of cultural and linguistic bias by the models.

\begin{table}
\centering
\setlength{\textfloatsep}{2pt}
\begin{tabular}{|l|c|c|}
\hline
\textbf{Model} & \textbf{VQA-UA} & \textbf{VQA-EN} \\ \hline
Qwen2-VL-7b & 34.41 & 83.12 \\ \hline
Paligemma-3b FT & 34.19 & 78.71 \\ \hline
Pixtral-12b & 31.55 & 46.32 \\ \hline
Gemini Pro & 32.73 & 40.15 \\ \hline
Paligemma-3b & 29.38 & 79.40 \\ \hline
GPT-4o & 29.04 & 40.69 \\ \hline
Llama3.2 & 27.57 & 42.41 \\ \hline
Claude Sonnet & 26.73 & 30.64 \\ \hline
\end{tabular}
\caption{Accuracy scores for VQA-1000-UA and VQA-1000-EN datasets, sorted by VQA-UA in descending order.}
\label{tab:vqa_results}

\end{table}

\label{ssec:instruction}
\noindent\textbf{Instruction-following Issues.}
The most prevalent challenge observed across models was inconsistent instruction following in Ukrainian. Even high-performing models like GPT-4o and Gemini frequently failed to respond in expected format. We have observed models replying in a much more verbose way than expected by Multi30K and VQA, therefore we modified prompts with an extra instruction to reply with one word for VQA or with a sentence for Multi30K, but issue persisted. 

\noindent\textbf{Code-switching issues.}
Besides instruction following, we've observed major issues with code-switching and language confusion.
This behavior was particularly pronounced in open-ended tasks like recipe generation and ingredient listing in the UACUISINE benchmark, where models would be prompted in Ukrainian but switch to English, Chinese or Russian for response. This suggests that current VLMs experience the same code-switching issues that are known to happen in text-only multilingual LLMs~\cite{borsh}. We have also observed the same issues of broken grammar \cyr{"Куряче супу з лапшой"}\latintext, non-existing words generation (\cyr{Хаширо-ітамэ, Курицики, Кулібино})\latintext and tokenization artifacts (\cyr{Рисotto}).\latintext

\noindent\textbf{Cultural misattribution.}
A key issue was cultural appropriation, notably when Ukrainian Borsch (UNESCO-recognized cultural heritage) was mislabeled as "Russian Red Borscht." This pattern extended to other Ukrainian dishes, with models defaulting to English or Russian translations even when prompted in Ukrainian. The misattribution went beyond labeling - in recipe generation, models often suggested Russian rather than traditional Ukrainian preparations. This systematic bias points to training data issues that risk reinforcing narratives diminishing Ukrainian cultural identity. Addressing this requires both improved Ukrainian language capabilities and better integration of accurate cultural knowledge in model training.

\noindent\textbf{Fine-tuning observations.}
The fine-tuned PaliGemma model demonstrated improvement in \autoref{tab:combined-uacuisine-results} on the UACUISINE benchmark compared to the base model, with exact match increasing from 3\% to 34\% and intersection match improving from 0\% to 15\%. This improvement suggests that even limited domain-specific fine-tuning can substantially enhance performance on culturally-focused tasks. Notably, this specialization also yields improved performance on the translated Ukrainian VQA benchmark (+5.33\%), while maintaining comparable performance on the English VQA benchmark (slight -0.52\% difference). These results suggest that targeted fine-tuning on Ukrainian cultural content can enhance model performance on Ukrainian language tasks without compromising general capabilities in English. We noticed a clear limitation of the VLMs to perform non-caption tasks that require more than general world knowledge, especially things like providing cuisine recipes, which manifests into low BERT score.

\begin{table}
\centering
\setlength{\textfloatsep}{2pt}
\begin{tabular}{|l|c|c|c|}
\hline
\textbf{Model} & \textbf{BERT} & \textbf{EM} & \textbf{IM} \\ 
\hline
Qwen2-VL-7B & 0.53 & 0.31 & 0.25 \\ 
Pixtral-12b & 0.55 & 0.02 & 0.00 \\ 
Llama3.2 & 0.57 & 0.01 & 0.25 \\  
Paligemma-3b   & 0.53 & 0.03 & 0.00 \\ 
\hline
Paligemma FT (1 ep) & 0.54 & 0.18 & 0.30 \\ 
Paligemma FT (2 ep) & 0.55 & 0.26 & 0.10 \\ 
Paligemma FT (4 ep) & 0.54 & 0.34 & 0.15 \\
\hline
\end{tabular}
\caption{UACUISINE Evaluation Metrics Across Models.}
\label{tab:combined-uacuisine-results}
\end{table}

\section{Conclusions}
In this work, we introduced a suite of benchmarks to evaluate VLMs in Ukrainian, addressing a critical gap in resources for low- and mid-resource languages. ZNO benchmark enables researchers to objectively estimate model performance for Ukrainian multimodal generation on expert-made questions. Our paper also reveals stark performance gaps between Ukrainian and English benchmarks as measured on VQA, underscoring the challenges faced by multimodal models in low-resource setting. 
We demonstrated the potential of targeted fine-tuning to improve cultural and linguistic relevance without sacrificing general capabilities. 
Future research directions should focus on extending benchmarks to include more diverse, language-specific tasks, addressing the culture gap. Beyond Ukrainian, the methodologies introduced here could serve as a template for advancing multimodal language modelling in underrepresented languages, enabling more inclusive access to AI instruments.

\section{Limitations} 
\label{sec:limitations}
While we believe that our work is a step forward in evaluating the Ukrainian capabilities of VLMs, it has a number of limitations. 

In particular, the usage of machine translation in translating the VQA dataset into the Ukrainian language might have introduced a bias, partially responsible for the degradation of performance comparing to the original VQA benchmank. 

Also, we have used the same prompt prefix for all queries to all the tested models. It is possible that this prompt might introduce a bias in model comparison.

ZNO dataset is heavily skewed towards STEM domains, having more than half of the questions in these categories. Also, STEM categories have the most visual-only questions (meaning that the model has to rely on its OCR capabilities to answer the question).

Multi30K provides both English and Ukrainian captions for the same image, which makes it suitable for testing multi-modal translation task. We haven't performed such testing.

\section{Ethical Considerations}
Several ethical considerations arise in developing and deploying multimodal benchmarks for Ukrainian language evaluation. Most critically, our work addresses questions of cultural representation and identity preservation, particularly salient given current geopolitical contexts. The systematic misattribution of Ukrainian cultural elements by AI models highlights risks of technological erasure of cultural identity. While our use of translated benchmarks enables comparative evaluation, this approach may inadvertently perpetuate biases and fail to capture uniquely Ukrainian contexts. Additionally, the observed tendency of models to default to Russian or English translations, even when prompted in Ukrainian, raises concerns about digital marginalization of Ukrainian language users. These considerations underscore the importance of developing culturally sensitive evaluation frameworks that can help ensure AI systems properly represent and serve Ukrainian language users.

\nocite{Ando2005,andrew2007scalable,rasooli-tetrault-2015}

\section*{Acknowledgments}

This research was made possible through generous support from several organizations. We thank \textbf{The alliance of De Novo and MK-Consulting} for providing computational resources, and \textbf{ELEKS} for their grant in memory of Oleksiy Skrypnyk. 

We also gratefully acknowledge \textbf{Amazon Web Services (AWS)} for cloud credits that enabled training and inference on H200 instances, and \textbf{Google Cloud Platform (GCP)} for credits supporting model training and inference.

\bibliography{custom}

\appendix
\newpage
\section{Evaluation breakdown by category}
\label{sec:appendix_A}
\addcontentsline{toc}{section}{Appendix A}

\begin{sidewaystable}[h]
\centering
\footnotesize 
\begin{tabular}{|p{0.15\textwidth}|c|c|c|c|c|c|c|c|c|c|c|c|c|}
\hline
\textbf{Model} & \textbf{Mathematics} & \textbf{Geography} & \textbf{Ukrainian} & \textbf{Other} & \textbf{Kindergarten} & \textbf{History} & \textbf{Spanish} & \textbf{German} & \textbf{French} & \textbf{English} & \textbf{Chemistry} & \textbf{Physics} & \textbf{Biology} \\ 
\hline
\textbf{Gemini Pro }                   & 0.489 & 0.687 & 0.800 & 0.705 & 0.661 & 0.722 & 1.000 & 0.600 & 0.900 & 0.957 & 0.732 & 0.612 & 0.767 \\ \hline
Claude 3.5 Sonnet & 0.541 & 0.623 & 0.722 & 0.500 & 0.590 & 0.643 & 0.714 & 0.600 & 0.931 & 0.963 & 0.725 & 0.510 & 0.560 \\ \hline
\textbf{Qwen2-VL-72B}      & 0.394 & 0.553 & 0.200 & 0.409 & 0.575 & 0.444 & 0.429 & 0.600 & 0.263 & 0.250 & 0.700 & 0.452 & 0.564 \\ \hline
GPT-4o                 & 0.609 & 0.260 & 0.100 & 0.182 & 0.195 & 0.167 & 0.143 & 0.000 & 0.100 & 0.171 & 0.747 & 0.550 & 0.218 \\ \hline
Qwen2-VL-7B      & 0.244 & 0.260 & 0.200 & 0.205 & 0.247 & 0.333 & 0.000 & 0.000 & 0.019 & 0.006 & 0.389 & 0.291 & 0.233 \\ \hline\hline
\textbf{Baseline} & 0.196 & 0.260 & 0.300 & 0.295 & 0.213 & 0.333 & 0.000 & 0.000 & 0.269 & 0.122 & 0.228 & 0.216 & 0.237 \\ \hline\hline
LLama-3.2-90B-Vision-Instruct & 0.0928 & 0.360 & 0.388 & 0.200 & 0.386 & 0.324 & 0.142 & 0.000 & 0.281 & 0.250 & 0.112 & 0.113 & 0.383 \\ \hline
Qwen2-VL-2B      & 0.151 & 0.050 & 0.000 & 0.000 & 0.006 & 0.000 & 0.000 & 0.000 & 0.000 & 0.000 & 0.247 & 0.200 & 0.019 \\ \hline
Llama-3.2-11B  & 0.062 & 0.150 & 0.000 & 0.159 & 0.124 & 0.111 & 0.143 & 0.000 & 0.106 & 0.213 & 0.061 & 0.066 & 0.154 \\ \hline
LLaVa-v1.6-mistral-7b & 0.061 & 0.100 & 0.000 & 0.068 & 0.052 & 0.056 & 0.000 & 0.000 & 0.006 & 0.037 & 0.087 & 0.078 & 0.049 \\ \hline
Paligemma-3b     & 0.084 & 0.003 & 0.000 & 0.000 & 0.000 & 0.000 & 0.000 & 0.000 & 0.000 & 0.000 & 0.114 & 0.079 & 0.008 \\ \hline
Pixtral-12b  & 0.000 & 0.000 & 0.000 & 0.000 & 0.000 & 0.000 & 0.000 & 0.000 & 0.000 & 0.000 & 0.000 & 0.000 & 0.000 \\ \hline
\end{tabular}
\caption{Performance metrics for ZNO test set, split by subject. The models highlighted are the best API model, Gemini, best open source model, Qwen2-VL-72B and Baseline - always select first choice. Notably, Claude 3.5 Sonnet demonstrates a similarly strong capabilities as Gemini.}
\label{tab:appendix_table_no_zno}
\end{sidewaystable}

\begin{sidewaystable}[h]
\centering
\footnotesize 
\begin{tabular}{|p{0.15\textwidth}|c|c|c|c|c|c|c|c|c|c|c|c|c|}
\hline
\textbf{Model} & \textbf{Mathematics} & \textbf{Geography} & \textbf{Ukrainian} & \textbf{Other} & \textbf{Kindergarten} & \textbf{History} & \textbf{Spanish} & \textbf{German} & \textbf{French} & \textbf{English} & \textbf{Chemistry} & \textbf{Physics} & \textbf{Biology} \\ 
\hline
\textbf{Gemini Pro }                     & 0.476 & 0.649 & 0.526 & 0.400 & 0.711 & 0.767 & 1.000 & 0.833 & 0.900 & 0.950 & 0.804 & 0.576 & 0.667 \\ \hline
Claude 3.5 Sonnet & 0.500 & 0.675 & 0.526 & 0.300 & 0.666 & 0.697 & 1.000 & 0.666 & 0.850 & 0.950 & 0.754 & 0.590 & 0.515 \\ \hline
\textbf{Qwen2-VL-72B}      & 0.366 & 0.514 & 0.368 & 0.200 & 0.356 & 0.651 & 0.429 & 0.500 & 0.300 & 0.100 & 0.745 & 0.470 & 0.606 \\ \hline
GPT-4o                 & 0.634 & 0.162 & 0.158 & 0.100 & 0.244 & 0.233 & 0.143 & 0.167 & 0.100 & 0.100 & 0.735 & 0.470 & 0.273 \\ \hline
LLama-3.2-90B-Vision-Instruct & 0.134 & 0.351 & 0.105 & 0.100 & 0.288 & 0.418 & 0.428 & 0.166 & 0.250 & 0.400 & 0.107 & 0.212 & 0.575 \\ \hline\hline
\textbf{Baseline}              & 0.232 & 0.162 & 0.263 & 0.000 & 0.289 & 0.349 & 0.143 & 0.167 & 0.250 & 0.150 & 0.225 & 0.182 & 0.212 \\ \hline\hline
Llama-3.2-11B & 0.049 & 0.189 & 0.105 & 0.100 & 0.111 & 0.140 & 0.000 & 0.167 & 0.200 & 0.250 & 0.098 & 0.121 & 0.121 \\ \hline
Qwen2-VL-2B      & 0.195 & 0.054 & 0.000 & 0.000 & 0.000 & 0.000 & 0.000 & 0.000 & 0.000 & 0.000 & 0.206 & 0.167 & 0.030 \\ \hline
LLaVa-v1.6-mistral-7b & 0.061 & 0.027 & 0.053 & 0.000 & 0.089 & 0.163 & 0.000 & 0.000 & 0.000 & 0.050 & 0.059 & 0.152 & 0.000 \\ \hline
Paligemma-3b    & 0.122 & 0.000 & 0.000 & 0.000 & 0.000 & 0.000 & 0.000 & 0.000 & 0.000 & 0.000 & 0.069 & 0.106 & 0.000 \\ \hline
Pixtral-12b  & 0.000 & 0.000 & 0.000 & 0.000 & 0.000 & 0.000 & 0.000 & 0.000 & 0.000 & 0.000 & 0.000 & 0.000 & 0.000 \\ \hline
\end{tabular}
\caption{Performance metrics for ZNO validation set, split by subject. The models highlighted are the best API model, Gemini, best open source model, Qwen2-VL-72B and Baseline - always select first choice. Notably, Claude 3.5 Sonnet demonstrates a similarly strong capabilities as Gemini.}
\label{tab:appendix_table_no_zno}
\end{sidewaystable}
\clearpage

\section{ZNO Dataset Overview}
\label{sec:appendix_B}

\begin{figure}[!h]
    \centering
    \includegraphics[width=\textwidth]{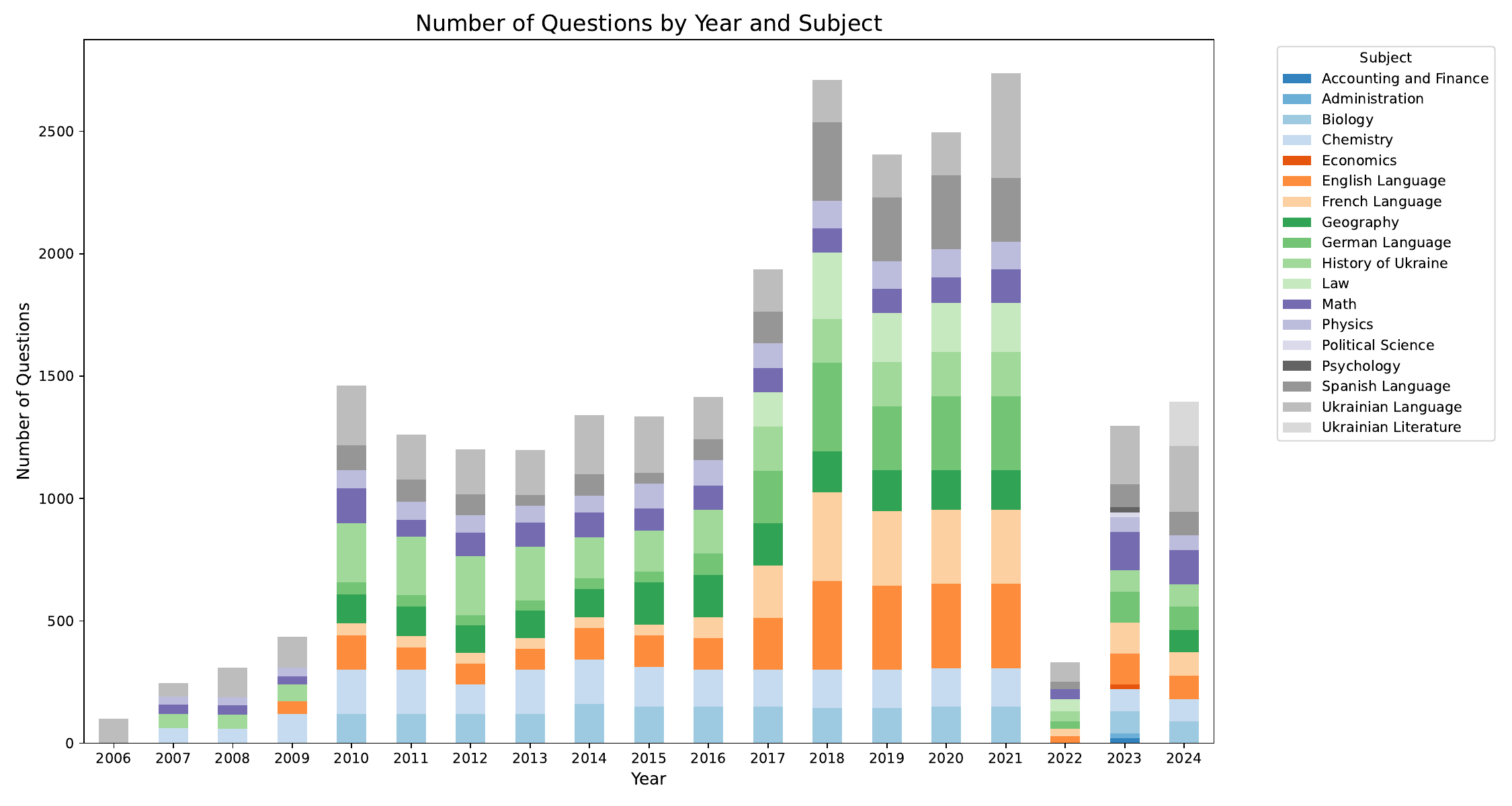}
    \caption{Distribution of ZNO questions by year and subject, showing a skewed yet diverse distribution of questions. }
    \label{fig:questions_by_year}
\end{figure}

\begin{table}[!h]
\centering
\begin{tabular}{lrrr}
\hline
\textbf{Category} & \textbf{Total} & \textbf{Visual-Only} & \textbf{Strictly-visual \%} \\
\hline

Chemistry & 1,021 & 946 & 92.65\% \\
Mathematics & 821 & 771 & 93.91\% \\
Physics & 661 & 595 & 90.02\% \\
History & 434 & 0 & 0.00\% \\
Geography & 374 & 0 & 0.00\% \\
Biology & 332 & 0 & 0.00\% \\
English language & 204 & 0 & 0.00\% \\
French language & 199 & 0 & 0.00\% \\
Kindergarten teaching & 134 & 0 & 0.00\% \\
Ukrainian language and literature & 56 & 0 & 0.00\% \\
Other & 31 & 0 & 0.00\% \\
Spanish language & 22 & 20 & 90.91\% \\
German language & 17 & 0 & 0.00\% \\
\hline
\end{tabular}
\caption{Distribution of ZNO questions by category. As we can see, STEM categories represent more than a half of a dataset, even having more than 90\% of all visual-only questions (typical question has a text and image, but in visual-only setting model hast to perform OCR to answer the question). }
\end{table}

\clearpage
\begin{table}[!h]
   \begin{tabular}{lrrr}
\hline
\textbf{Subset} & \textbf{\# Questions} & \textbf{Visual-Only} & \textbf{Visual \%} \\
\hline
Dev & 491 & 235 & 47.86\% \\
Validation & 490 & 233 & 47.55\% \\
Test & 3,325 & 1,864 & 56.06\% \\
\hline
\textbf{Total} & 4306 & 2332 & 54.16\% \\
\hline
\end{tabular}
\caption{Distribution of ZNO Dataset by subset. Dev and Validation subset each represent 10\% of all data that can be used during model training.}

\end{table}

\section{UACUISINE Metrics}
\label{sec:intersection-match}
For the dish name prediction and characteristic classification we use exact match score (EM):
    \begin{equation}
        \text{EM}(y, \hat{y}) = \mathbb{1}(y = \hat{y}) 
    \end{equation}
    where:
    \begin{itemize}
        \item EM is the exact match score
        \item $y$ is the ground truth
        \item $\hat{y}$ is the model prediction
    \end{itemize}
    This score is calculated on the stem level to account for gender based noun changes in Ukrainian.

For the ingredients generation we use intersection match score (IM):
    \begin{equation}
        \text{IM}(y, \hat{y}) = \frac{|\mathcal{W}(y) \cap \mathcal{W}(\hat{y})|}{|\mathcal{W}(y) \cup \mathcal{W}(\hat{y})|}
    \end{equation}
    where:
    \begin{itemize}
        \item IM is the intersection match score
        \item $\mathcal{W}(\cdot)$ represents the set of words in a string
    \end{itemize}

\clearpage
\section{UACUISINE Questions}
\label{sec:appendix_C}
\addcontentsline{toc}{section}{Appendix C}

\begin{table}[h]
\centering
\begin{tabular}{|p{0.5\textwidth}|p{0.5\textwidth}|} 
\hline
\textbf{Ukrainian Question} & \textbf{English Translation} \\
\hline
\multicolumn{2}{|c|}{\textit{1. Identification Questions}} \\
\hline
\begin{otherlanguage}{ukrainian}Q1: Як називається ця страва?\end{otherlanguage} & Q1: What is this dish called? \\
\hline
\begin{otherlanguage}{ukrainian}Q2: Яка назва цієї страви в Україні?\end{otherlanguage} & Q2: What is the name of this dish in Ukraine? \\
\hline
\begin{otherlanguage}{ukrainian}Q3: Я обідаю в українському ресторані. Зараз я збираюся їсти цю страву. Як називається ця страва?\end{otherlanguage} & Q3: I'm having lunch at a Ukrainian restaurant. I'm about to eat this dish. What is this dish called? \\
\hline
\multicolumn{2}{|c|}{\textit{2. Generation Questions}} \\
\hline
\begin{otherlanguage}{ukrainian}Q4: Перелічи інгредієнти необхідні для приготування зображеної страви\end{otherlanguage} & Q4: List the ingredients needed to prepare the shown dish \\
\hline
\begin{otherlanguage}{ukrainian}Q5: Як приготувати цю страву. Опиши коротко покроковий рецепт.\end{otherlanguage} & Q5: How to prepare this dish. Describe briefly the step-by-step recipe. \\
\hline
\multicolumn{2}{|c|}{\textit{3. Binary Classification Questions}} \\
\hline
\begin{otherlanguage}{ukrainian}Q6: Яка ця страва на смак: солона чи солодка?\end{otherlanguage} & Q6: How does this dish taste: salty or sweet? \\
\hline
\begin{otherlanguage}{ukrainian}Q7: Ця страва подається холодна чи гаряча?\end{otherlanguage} & Q7: Is this dish served cold or hot? \\
\hline
\end{tabular}
\caption{Ukrainian UACUISINE Questions with English Translations}
\label{tab:ukrainian-questions}
\end{table}

\end{document}